\documentclass{article}

\usepackage{iclr2026_conference,times}

\iclrfinalcopy

\usepackage[utf8]{inputenc} 
\usepackage[T1]{fontenc}    
\usepackage[pagebackref, breaklinks, colorlinks]{hyperref} 
\definecolor{mylinkcolor}{rgb}{0.72,0.10,0.10}  
\definecolor{mycitecolor}{rgb}{0.0,0.50,0.30}   
\definecolor{myurlcolor}{rgb}{0.10,0.32,0.72}   
\definecolor{Terracotta}{RGB}{204,78,92}         
\hypersetup{linktoc=page, linkcolor=mylinkcolor, citecolor=mycitecolor, urlcolor=myurlcolor}
\usepackage{url}            
\usepackage{booktabs}       
\usepackage{amsfonts}       
\usepackage{nicefrac}       
\usepackage{microtype}      
\usepackage{xcolor}         
\usepackage{graphicx}       
\usepackage{wrapfig}        
\usepackage{xspace}         
\usepackage{algorithm}      
\usepackage{algpseudocode} 
\usepackage{multirow}       

\makeatletter
\newif\ifappendixtoc
\let\origappendix\appendix
\let\origaddcontentsline\addcontentsline
\renewcommand{\appendix}{%
  \origappendix
  \appendixtoctrue
}
\renewcommand{\addcontentsline}[3]{%
  \origaddcontentsline{#1}{#2}{#3}%
  \ifappendixtoc
    \def\appendixtocfile{#1}%
    \def\maintocfile{toc}%
    \ifx\appendixtocfile\maintocfile
      \origaddcontentsline{apc}{#2}{#3}%
    \fi
  \fi
}
\newcommand{\appendixsectiontocline}[2]{%
  \addpenalty{\@secpenalty}%
  \addvspace{1.0em \@plus\p@}%
  \@tempdima 1.5em%
  \begingroup
    \parindent \z@
    \rightskip \@pnumwidth
    \parfillskip -\@pnumwidth
    \leavevmode\bfseries
    \advance\leftskip\@tempdima
    \hskip -\leftskip
    #1\nobreak
    \leaders\hbox{$\m@th\mkern\@dotsep mu.\mkern\@dotsep mu$}\hfill
    \nobreak\hb@xt@\@pnumwidth{\hss #2}\par
  \endgroup
}
\newcommand{\appendixnavigation}{%
  \section*{Appendix}%
  \begingroup
    \let\l@section\appendixsectiontocline
    \renewcommand*\l@subsection{\@dottedtocline{2}{1.5em}{2.3em}}%
    \renewcommand*\l@subsubsection{\@dottedtocline{3}{3.8em}{3.2em}}%
    \@starttoc{apc}%
  \endgroup
  \clearpage
}
\makeatother

\algrenewcommand\algorithmiccomment[1]{\hfill$\triangleright$ #1}

\newcommand{\ourmodel}{WorldWeaver\xspace}
\newcommand{\ourmodelshort}{\ensuremath{\mathbf{W}^{\mathbf{2}}}\xspace}
\newcommand{\eg}{\textit{e.g.}\xspace}

\title{Streaming Multi-Agent Autoregressive Diffusion Model with World State Registers}

\author{%
Sicheng Mo\textsuperscript{*~1} \quad
Yuheng Li\textsuperscript{*~2} \quad
Ziyang Leng\textsuperscript{1} \quad
Krishna Kumar Singh\textsuperscript{2} \quad
Bolei Zhou\textsuperscript{1} \\
\\
\textsuperscript{1}University of California, Los Angeles \quad
\textsuperscript{2}Adobe Research \\
}

\begin{document}

\maketitle
\lhead{Preprint.}
\begingroup
\renewcommand{\thefootnote}{\fnsymbol{footnote}}%
\footnotetext[1]{Equal contribution.}%
\endgroup

\begin{figure}[h]
\centering
\makebox[\textwidth][c]{%
  \includegraphics[width=1.3\linewidth]{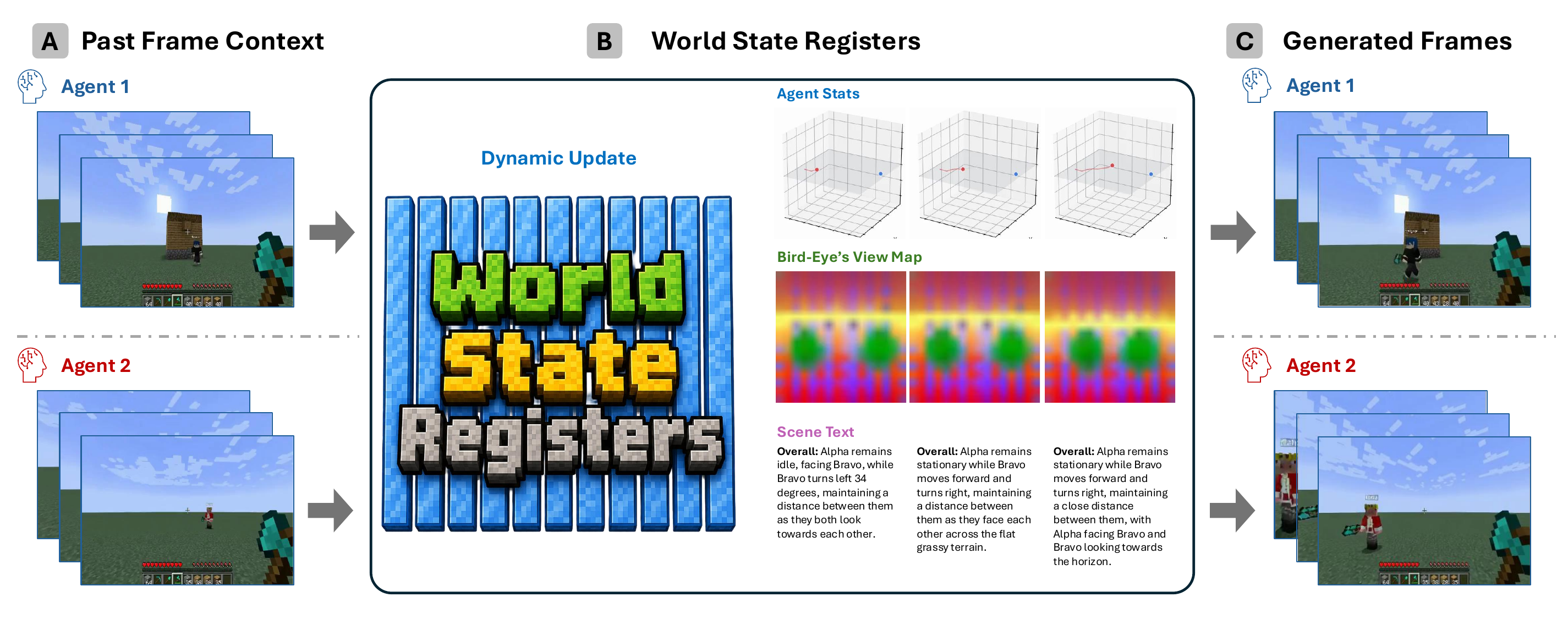}%
}
\vspace{-1.0em}
\caption{Our model enables interactive agents acting in a shared world. During rollout, a set of register tokens represents the evolving world state shared across agents. This state captures agent status, bird's-eye views, and scene text, and is being dynamically updated at each rollout step while new frames are generated.}
\label{fig:teaser}
\end{figure}

\begin{abstract}
Multi-agent interactive world models should not only generate consistent observations, but also maintain world states that persist across agents and evolve across views.
Existing autoregressive video diffusion pipelines carry forward observation history as conditioning context, which makes shared state difficult to maintain in multi-agent and multi-view settings.
We present \ourmodel{} (\ourmodelshort{}), a streaming multi-agent video diffusion model that augments rollout with cross-agent world state registers: learnable tokens that store shared world information, track individual agent status, and are dynamically updated after each generated chunk.
We ground these registers with supervision signals spanning individual agent status, global state views including bird's-eye views, and scene text.
We further improve the architecture with a Mixture-of-Transformers design that uses separate weights for world state modeling and visual frame modeling.
Extensive experiments in two-agent Minecraft video generation show that explicit world-state modeling improves logical consistency and generation quality.
Project Page: \url{https://vail-ucla.github.io/worldweaver/}
\end{abstract}

\section{Introduction}
\label{sec:introduction}

Recent advances in generative modeling have enabled increasingly powerful interactive and streaming world models (WMs)~\citep{openaiSoraCreating,savva2026solaris}.
Most video world models simulate a scene from the view of a single agent.
While this design can produce visually plausible frames, it is challenging to maintain logical and geometric consistency.
This issue is further amplified in multi-agent settings~\citep{savva2026solaris}, where each observer only sees a partial 2D projection of the same underlying 3D world, while observations across agents must remain compatible with that shared state.
Beyond these cross-agent constraints, the state of the world continues to evolve no matter a particular observer sees it or not.
This perspective motivates us to model the underlying world state directly, rather than treating consistency as a byproduct of per-agent video generation.

Temporally consistent video can be generated by chunk-based autoregressive video diffusion models~\citep{chen2024diffusionforcing,yin2024causvid}.
In such a pipeline, past frames are treated as conditioning context, and the model generates the next video chunk with joint spatiotemporal attention.
However, because each chunk must be decoded into pixel space, the model needs to re-infer world information at every generation step.
As a result, it carries forward limited observation history rather than explicit world states that can account for both observed and unobserved changes.
In multi-view settings, this further introduces a mismatch between each agent's observation and the shared world state.

To tackle this issue, we explicitly model world states inside the generation process, instead of leaving them to be repeatedly inferred from frame history.
We introduce a world state register, a group of learnable tokens that captures shared world information and is being incrementally updated during generation.
The register is designed to carry persistent global information and individual agent status.
We define two key properties for the world state register: (1) being persistent across agents and rollout steps, and (2) being dynamically updateable when new observations arrive.
In this way, the observations from all agents contribute to the world state update, encouraging temporal and cross-agent consistency at the level of the underlying world state.

To facilitate long-horizon video generation, we adapt the Self-Forcing paradigm~\citep{huang2025selfforcing} to update both frames and world registers during rollout.
After each denoised video chunk, the model refreshes the world register from the previous register and the newly generated observations, so the updated register can condition the next chunk.
We further ground the register with complementary supervision signals: agent status encourages local motion consistency, bird's-eye views constrain global geometry, and scene text provides semantic grounding.
We also explore the effect of semi-supervised world-state training, where a small labeled subset can anchor register semantics while additional unlabeled videos continue to improve the generative rollout.

We also study architectures for world-state modeling in diffusion models.
When dense transformer weights must generate pixels and update world registers, the frame and state objectives can compete, especially once the register is pushed toward richer supervised semantics.
We therefore use a Mixture-of-Transformers (MoT) architecture~\citep{liang2024mixtureoftransformers} that treats the world register branch as a distinct state pathway while keeping it coupled to the visual rollout.
This separation improves supervised register learning and stabilizes frame register self-forcing during streaming generation.

We summarize our contributions as follows:
\begin{enumerate}
    \item We introduce a new design of world state registers, a persistent and dynamically updateable set of tokens for carrying shared world information across agents and rollout steps.
    \item We explore the supervision space for grounding world registers, ranging from individual agent status to global world-state signals such as bird's-eye views and scene text.
    \item We propose an improved MoT-based architecture for video diffusion models that supports joint generation of world states and video frames. Our experiments show that this design improves performance and reduces conflicts between state updating and frame generation.
\end{enumerate}

\section{Related Work} 
\label{sec:related_works}

\smallskip
\noindent
\textbf{Streaming video diffusion models.}
Large-scale diffusion models provide a flexible foundation for image and video generation. Video diffusion models~\citep{openaiSoraCreating,ma2024latte,wang2023lavie} use bidirectional attention to synthesize coherent visual content and smooth temporal transitions under a shared text condition. Building on these backbones, autoregressive video diffusion methods~\citep{chen2024diffusionforcing,yin2024causvid} introduce chunk level causal attention and KV caching to support streaming generation. Self-Forcing~\citep{huang2025selfforcing} further reduces the train test mismatch by rolling out the model on its own generated frames during training, while recent extensions~\citep{shin2025motionstream,cui2025selfforcingplus,zhu2026causalforcing,zhao2026relaxforcing} improve minute level streaming with attention sinks and stronger positional encoding. These advances make long horizon generation increasingly practical, but their main objective remains generation quality and rollout stability. In contrast, our \ourmodel{} builds on streaming video diffusion for interactive generation and uses persistent world state registers to generate a logically consistent world, where temporal and cross agent consistency are grounded in an explicit state beyond the local context window.

\smallskip
\noindent
\textbf{Memory-based video generation.} 
Memory mechanisms have been widely explored in video generation to preserve information that is not fully specified by the current frame or local context, including appearance, identity, geometry, object states, and scene layout. Explicit memory methods introduce retrieval buffers, spatial maps, 3D representations, or 4D scene memories that the generator can consult~\cite{xiao2026worldmem,yu2025contextasmemory,yu2026mosaicmem}. This design is closely related to 3D novel view synthesis because both use a queryable scene representation to preserve geometry, appearance, and layout across viewpoints~\cite{team2026inspatioworld,yang2026neoverse,sun2025worldplay}. Existing explicit modeling, however, remains limited by scalability and by the need to update dynamic object states over time. Implicit memory methods store history inside the generative model through past-frame conditioning, recurrent rollout, or attention KV caches~\cite{chen2024diffusionforcing,yin2024causvid,huang2025selfforcing,zhao2026relaxforcing,gao2026lome}, but they provide weaker constraints because the memory is entangled with visual tokens and biased toward recent observations.
Another line of work also values the role of state modeling in generative videos to ensure logical correctness~\citep{gammaworld2026,po2025long,chen2025recurrent,po2026multigen}.
In contrast, our method keeps the memory inside the generator as persistent world state registers, but trains it as implicit memory with explicit supervision from individual agent states and global scene states, so the stored state receives direct grounding for world generation beyond frame history alone.

\smallskip
\noindent
\textbf{World understanding and generation in unified models.}
Visual generation and understanding are not fully decoupled. Text-to-image diffusion models have shown useful visual representations for discriminative tasks such as classification~\citep{li2023yourdiffclassifier,clark2023zeroshotclassifier}, detection~\citep{xu20243difftection}, and segmentation~\citep{xu2023odise,tian2024diffseg,namekata2024emerdiff}. A line of unified multimodal models further makes this connection explicit by training a single model for both text-to-image generation and image-to-text understanding~\citep{zhou2024transfusion,emu3,xie2024show}. Another line of work~\citep{mo2025xfusion,shi2024llamafusion,deng2025bagel,yang2026omni}, including Mixture-of-Transformers~\citep{liang2024mixtureoftransformers}, explores the use of separate weights or dedicated fusion modules for different modalities and token roles, extending unified multimodal pretraining to broader interleaved and context rich settings. Meanwhile, MetaMorph~\citep{tong2025metamorph} and X-Fusion~\citep{mo2025xfusion} suggest that understanding tasks can improve generation by shaping stronger multimodal representations.
Our \ourmodel{} builds on these observations and further explores the synergy between world understanding and generation through world state registers. These registers provide a structured and verifiable representation with additional grounding from individual agent states and global scene states.

\section{Method}
\label{sec:method}

\subsection{Overview}

We adopt the multi-agent world-model setting of Solaris~\citep{savva2026solaris}. Multiple agents act in a shared world, and the model predicts each agent's future video stream from past observations and actions. We extend a single-player latent video diffusion pipeline by attaching a player axis to every generated tensor. This gives a compact tensor setting for synchronized multi-agent generation. With \(P\) agents, the per-step observation and action are \(\mathbf{x}^{t}=(\mathbf{x}^{t}_{1},\ldots,\mathbf{x}^{t}_{P})\) and \(\mathbf{a}^{t}=(\mathbf{a}^{t}_{1},\ldots,\mathbf{a}^{t}_{P})\). A clip of \(N\) latent frames forms \(\mathbf{x}\in\mathbb{R}^{P\times N\times H\times W\times C}\). The conditioning \(c\) bundles a first-frame visual embedding, a masked first-frame latent, and each agent's action sequence, with concrete dimensions deferred to the experiments. We introduce the stage-specific objectives in the subsections below.

We next introduce \emph{world state registers} as a separate cross-agent state pathway. These registers are learnable tokens that capture hidden information about the shared world. They are shared across agents and updated incrementally as new frames are generated. The registers live alongside frame tokens in a causal transformer, so a single forward pass can denoise the next frame and refresh the cross-agent world state at the same time.
We further use a Mixture-of-Transformers (MoT) design that assigns separate weights to register and frame tokens while keeping joint attention over the interleaved sequence. Detailed implementation is provided in Appendix~\ref{app:mot-implementation}.

Our model follows a three-stage training curriculum on top of standard single-agent pretraining.
\emph{Stage 1 (Bidirectional Training)} finetunes a multi-player teacher on synchronized multi-agent clips. The teacher has full temporal context and learns cross-agent scene structure with bidirectional attention.
\emph{Stage 2 (Causal Training)} converts this teacher into a causal student by continuing the flow matching objective with causal attention and register supervision.
\emph{Stage 3 (Self-Forcing)} rolls out the student on its own predictions. After each generated frame, the model commits the updated register, so register drift is exposed together with frame drift. This stage uses DMD style distribution matching from the bidirectional teacher while retaining state supervision.

\begin{figure*}[t]
\centering
\includegraphics[width=\textwidth]{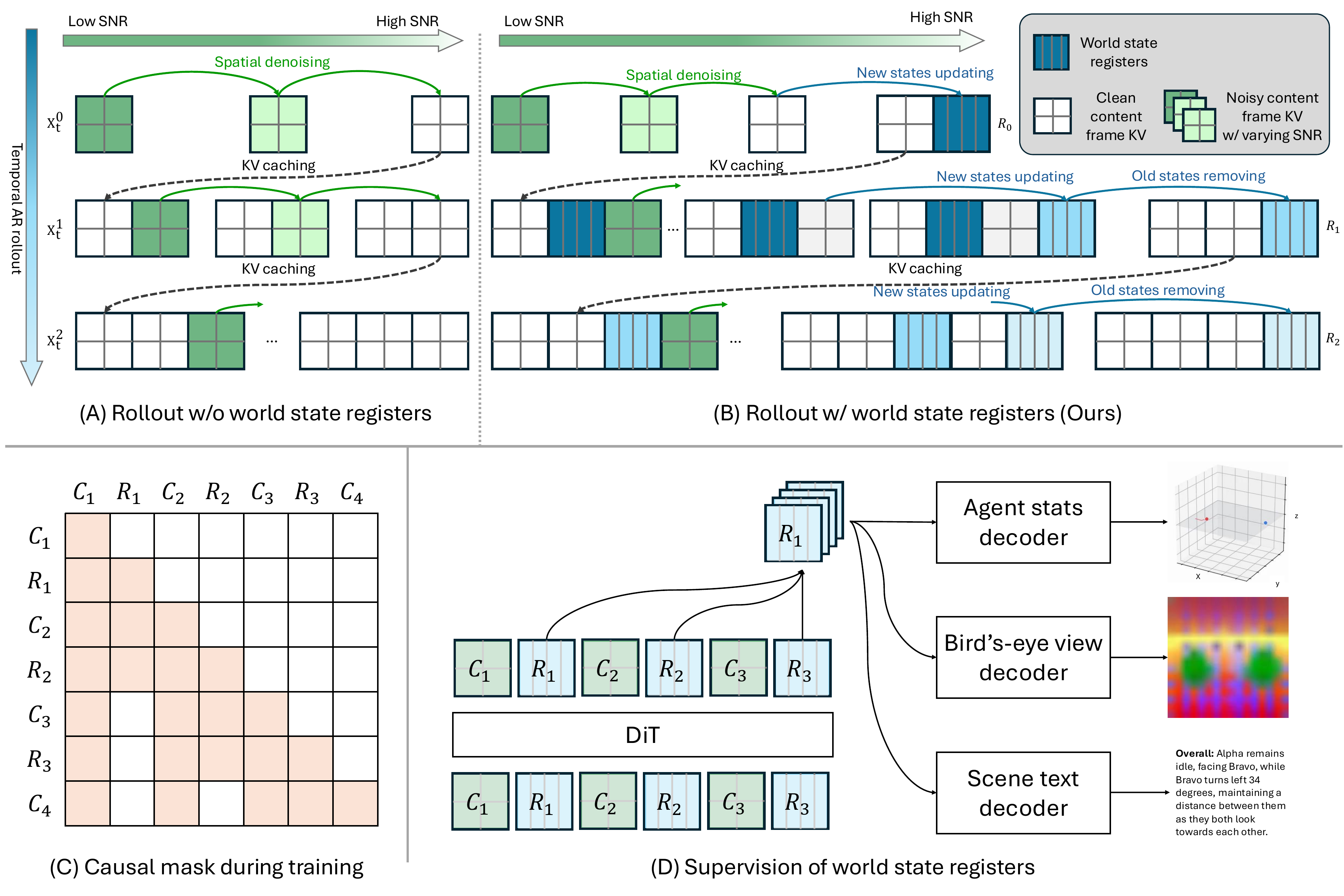}
\vspace{-2.0em}
\caption{Pipeline overview. (A)~A standard streaming autoregressive diffusion model denoises each new frame from a local KV cache window. (B)~Our model streams with world state registers, which carry global state information and individual agent status. (C)~Causal attention mask during Stage-2 training. Frame tokens attend to the local window and the latest register. (D)~Auxiliary decoders ground the registers with agent status, bird's-eye-view maps, and scene text, encouraging the world state registers to encode global scene information.}
\label{fig:pipeline}
\vspace{-1.0em}
\end{figure*}

\subsection{Stage 1: Bidirectional Training}

 Stage 1 trains a multi-player bidirectional teacher that jointly denoises the latent sequences of all players under bidirectional and cross-player attention. This teacher provides a full-time estimation of synchronized scene structure, including shared layout, relative player motion, and cross-view consistency. We initialize it from the pretrained single-player model to preserve the visual generation prior while adapting the model to synchronized multi-player data. We train this teacher with the conditional flow matching objective that predicts the velocity field \(\mathbf{v}_{\theta}(\mathbf{x}_t, c, t)\):
\begin{equation}
\mathcal{L}_{\mathrm{flow}}
=
\mathbb{E}
\left[
\left\|
\mathbf{v}_{\theta}(\mathbf{x}_t, c, t) - (\epsilon - \mathbf{x}_0)
\right\|_2^2
\right].
\end{equation}

\subsection{Stage 2: Causal Training with World State Registers}
\label{sec:causal-world-cache}

In this stage, we initialize the causal model from the bidirectional Stage 1 checkpoint and continue training with the same flow matching loss defined above.
Following Solaris~\citep{savva2026solaris} and Diffusion Forcing~\citep{chen2024diffusionforcing}, we sample an independent noise level for each player and latent frame, apply a block-level causal mask, and maintain a sliding window of only the latest \(W\) past latent frames.
This training gives the student autoregressive generation capability, where each new frame is conditioned on the action input and past frames. However, this design does not ground world modeling beyond the local context window, and therefore can fail to preserve a logically consistent world shared across agents.

To address this issue, we explicitly model the world state rather than relying on local frame history alone.
The world state represents persistent information at a specific time step and summarizes the shared scene, including global layout, object configuration, and agent status that may not be visible from every player's current view.
By carrying this state across rollout steps, the world model can synchronize predictions across agents and preserve logically consistent cross-agent interactions.

In practice, we define the world state registers (\textbf{WSR}) as a group of learnable \(K\) tokens \(\mathbf{r}_i\in\mathbb{R}^{K\times d}\).
We formally define the streaming mechanism of world models with WSR as follows:
\[
\mathbf{r}_i=G_\theta(\mathbf{r}_{i-1},\mathbf{x}_{i-W+1},\ldots,\mathbf{x}_i,a_i),\qquad p_\theta(\mathbf{x}_{i+1} \mid \mathbf{x}_{i-W+1},\ldots,\mathbf{x}_i, a_{i+1},\mathbf{r}_i).
\]
During rollout, each register \(\mathbf{r}_i\) is committed after the context frame \(\mathbf{x}_i\) has been generated, thereby serving as the state summary through frame \(i\) that conditions the next context frame.

Inspired by causal masking in Diffusion Forcing~\citep{chen2024diffusionforcing}, we interleave frame tokens and register groups as \([\mathbf{x}_0,\mathbf{r}_0,\mathbf{x}_1,\mathbf{r}_1,\ldots,\mathbf{x}_{F-1},\mathbf{r}_{F-1}]\).
This ordering makes the rollout causal at the state level: the committed register precedes the next frame and the frame is conditioned on that register.
Concretely, \(\mathbf{r}_i\) precedes and conditions \(\mathbf{x}_{i+1}\), so state update and frame prediction alternate across the sequence.
Combined with the local attention window of size \(W\), each register query attends to the same local context frames ending at its commit step, along with the immediately preceding register.
A register query for \(\mathbf{r}_i\) thus attends to \([\mathbf{x}_{i-W+1},\ldots,\mathbf{x}_i]\) and \(\mathbf{r}_{i-1}\), as shown in Figure~\ref{fig:pipeline}(C).


Stage 2 keeps the flow matching loss and register supervision within a single training objective, where the flow matching term gives the causal student frame-level supervision under the committed register.
However, this signal alone does not specify what information the WSR pathway should retain.
We therefore ground each committed register with auxiliary world state predictions, as shown in Figure~\ref{fig:pipeline}(D).
For each supervision type \(m\in\mathcal{M}\), the prediction head output \(h_m(\mathbf{r}_i)\), the same step target \(\mathbf{y}_i^m\), and the metric \(d_m\) make the prediction, target, and distance calculation explicit.
In our experiments, we use three supervision signals:
\begin{enumerate}
\item \textbf{Agent status.} The head predicts a simulator state vector \(\hat{\mathbf{y}}_i^{\mathrm{agent}}=h_{\mathrm{agent}}(\mathbf{r}_i)\), and the target \(\mathbf{y}_i^{\mathrm{agent}}\) contains position, velocity, and orientation. We use \(d_{\mathrm{agent}}=\|\hat{\mathbf{y}}_i^{\mathrm{agent}}-\mathbf{y}_i^{\mathrm{agent}}\|_2^2\).
\item \textbf{Bird's-eye-view map.} The head predicts a bird's-eye view \(\hat{\mathbf{y}}_i^{\mathrm{bev}}=h_{\mathrm{bev}}(\mathbf{r}_i)\), supervised by the aligned bird's-eye target \(\mathbf{y}_i^{\mathrm{bev}}\). We use \(d_{\mathrm{bev}}=1-\cos(\hat{\mathbf{y}}_i^{\mathrm{bev}},\phi_{\mathrm{DINOv2}}(\mathbf{y}_i^{\mathrm{bev}}))\), where \(\phi_{\mathrm{DINOv2}}\) denotes a frozen DINOv2 encoder~\citep{oquab2023dinov2}.
\item \textbf{Scene text.} The head predicts token logits \(\hat{\mathbf{y}}_i^{\mathrm{text}}=h_{\mathrm{text}}(\mathbf{r}_i)\), and the target \(\mathbf{y}_i^{\mathrm{text}}\) is a textual scene description. We use \(d_{\mathrm{text}}=\mathrm{CE}(\hat{\mathbf{y}}_i^{\mathrm{text}},\mathbf{y}_i^{\mathrm{text}})\).
\end{enumerate}
Together with the flow matching loss, we train the student with the following objective:
\[
\mathcal{L}_{\mathrm{reg}}=\sum_{i,m}\lambda_m d_m\left(h_m(\mathbf{r}_i),\mathbf{y}_i^m\right),\qquad \mathcal{L}_{\mathrm{S2}}=\mathcal{L}_{\mathrm{flow}}+\lambda_{\mathrm{state}}\mathcal{L}_{\mathrm{reg}}.
\]
Here the sum ranges over rollout steps and the three target types, and we set \(\lambda_{\mathrm{state}}\) to one by default in Stage 2.
In the Stage 2 objective, \(\mathcal{L}_{\mathrm{flow}}\) remains the frame-generation flow matching loss, while \(\mathcal{L}_{\mathrm{reg}}\) assigns explicit state targets to the committed registers.
The prediction heads are training-only modules and are discarded at inference, so this supervision does not change the rollout cost.
We ablate these supervision types in Section~\ref{sec:supervision-ablation}.

\begin{figure*}[t]
\centering
\small
\begin{minipage}[t]{0.48\textwidth}
\vspace{0pt}
\hrule
\smallskip
\refstepcounter{algorithm}\label{alg:world-cache-rollout}
\textbf{Algorithm \thealgorithm} Autoregressive Diffusion Inference\\
with World State Registers
\smallskip
\hrule
\smallskip
\begin{algorithmic}[1]
\Require Frame KV cache size \(W\)
\Require Conditioning \(C\)
\Require Denoising time steps \(\{t_1,\ldots,t_T\}\)
\Require Number of generated frames \(M\)
\Require Causal AR diffusion model \(G_\theta\), returning KV via \(G_\theta\)
\State \textbf{Initialize} generated latents \(\mathbf{X}_\theta\gets[\,]\)
\State \textbf{Initialize} frame KV cache \(\mathrm{KV}_{C}\gets[\,]\)
\State \textbf{Initialize} register KV cache \(\mathrm{KV}_{R}\gets[\,]\)
\State \textbf{Initialize} world state \(\mathbf{r}_0\)
\State \(\mathrm{KV}_{R}.\mathrm{append}(\mathbf{r}_0)\)
\For{\(i=1,\ldots,M\)}
    \State \textbf{Initialize} \(\mathbf{x}^i_{t_T}\sim\mathcal{N}(0,I)\)
    \For{\(j=T,\ldots,1\)}
        \State Set \(\mathbf{KV}\gets(\mathrm{KV}_{C},\mathrm{KV}_{R})\)
        \State Set \(\hat{\mathbf{x}}^i_0\gets G_\theta(\mathbf{x}^i_{t_j};t_j,\mathbf{KV},c_i)\)
        \If{\(j=1\)}
            \State \(\mathbf{X}_\theta.\mathrm{append}(\hat{\mathbf{x}}^i_0)\)
            \State Set \(\mathrm{KV}_{C}^{i}\gets G_\theta(\hat{\mathbf{x}}^i_0;0,\mathbf{KV},c_i)\)
            \If{\(|\mathrm{KV}_{C}|=W\)}
                \State \(\mathrm{KV}_{C}.\mathrm{pop}(0)\)
            \EndIf
            \State \(\mathrm{KV}_{C}.\mathrm{append}(\mathrm{KV}_{C}^{i})\)
            \State Set \(\mathbf{r}_{i}\gets G_\theta(\mathrm{KV}_{R},\hat{\mathbf{x}}^i_0,c_i)\)
            \State Set \(\mathrm{KV}_{R}\gets[\mathbf{r}_0,\mathbf{r}_{i}]\)
        \Else
            \State sample \(\epsilon\sim\mathcal{N}(0,I)\)
            \State Set \(\mathbf{x}^i_{t_{j-1}}\gets\Psi(\hat{\mathbf{x}}^i_0,\epsilon,t_{j-1})\)
        \EndIf
    \EndFor
\EndFor
\State \textbf{return} \(\mathbf{X}_\theta\)
\end{algorithmic}
\smallskip
\hrule
\end{minipage}\hfill
\begin{minipage}[t]{0.48\textwidth}
\vspace{0pt}
\hrule
\smallskip
\refstepcounter{algorithm}\label{alg:sf-world-cache}
\textbf{Algorithm \thealgorithm} Self-Forcing Training with\\
Frame and State Register Rollout
\smallskip
\hrule
\smallskip
\begin{algorithmic}[1]
\Require Conditioning \(C\)
\Require Denoising time steps \(\{t_1,\ldots,t_T\}\)
\Require Number of video frames \(N\)
\Require Causal AR diffusion model \(G_\theta\), returning KV via \(G_\theta\)
\Loop
\State \textbf{Initialize} generated latents \(\mathbf{X}_\theta\gets[\,]\)
\State \textbf{Initialize} frame KV cache \(\mathrm{KV}_{C}\gets[\,]\)
\State \textbf{Initialize} register KV cache \(\mathrm{KV}_{R}\gets[\,]\)
\State \textbf{Initialize} world state \(\mathbf{r}_0\)
\State \(\mathrm{KV}_{R}.\mathrm{append}(\mathbf{r}_0)\)
\State sample \(s\sim\mathrm{Uniform}(1,\ldots,T)\)
\For{\(i=1,\ldots,N\)}
    \State \textbf{Initialize} \(\mathbf{x}^i_{t_T}\sim\mathcal{N}(0,I)\)
    \For{\(j=T,\ldots,s\)}
        \State Set \(\mathbf{KV}\gets(\mathrm{KV}_{C},\mathrm{KV}_{R})\)
        \State Set \(\hat{\mathbf{x}}^i_0\gets G_\theta(\mathbf{x}^i_{t_j};t_j,\mathbf{KV},c_i)\)
        \If{\(j=s\)}
            \State \(\mathbf{X}_\theta.\mathrm{append}(\hat{\mathbf{x}}^i_0)\)
            \State Set \(\mathrm{KV}_{C}^{i}\gets G_\theta(\hat{\mathbf{x}}^i_0;0,\mathbf{KV},c_i)\)
            \State \(\mathrm{KV}_{C}.\mathrm{append}(\mathrm{KV}_{C}^{i})\)
            \State Set \(\mathbf{r}_{i}\gets G_\theta(0,\mathbf{KV},c_i)\)
            \State Set \(\mathrm{KV}_{R}\gets[\mathbf{r}_0,\mathbf{r}_{i}]\)
        \Else
            \State sample \(\epsilon\sim\mathcal{N}(0,I)\)
            \State Set \(\mathbf{x}^i_{t_{j-1}}\gets\Psi(\hat{\mathbf{x}}^i_0,\epsilon,t_{j-1})\)
        \EndIf
    \EndFor
\EndFor
\State Update \(\theta\) with \(\mathcal{L}_{\mathrm{S3}}\), and update \(s_{\mathrm{fake}}\) on student rollouts
\EndLoop
\end{algorithmic}
\smallskip
\hrule
\end{minipage}
\end{figure*}

\subsection{Stage 3: Self-Forcing with Context Frame and State Rollout}

We define the inference rollout as the autoregressive procedure used by our world model at deployment.
At each step, the model denoises the next latent frame from the current frame KV cache, the latest committed register, and the action input.
It then appends the generated frame to the cache and refreshes the world state register to summarize the updated world state.
Algorithm~\ref{alg:world-cache-rollout} gives this inference rollout, and Figure~\ref{fig:pipeline}(B) visualizes the same frame register loop.

However, the Stage 2 model is not yet ready for long-horizon rollout.
Teacher forcing trains the student on ground truth causal context, while inference uses its own generated frames and committed registers, allowing frame and state errors to propagate across rollout.
Following Self-Forcing, Stage 3 addresses the mismatch by simulating this rollout during training and exposing the causal student to its own generated history.
We define the self-forcing training loop as a rollout that first runs the student on generated context frames and committed registers, then applies the self-forcing objective to the resulting clean prediction. Algorithm~\ref{alg:sf-world-cache} summarizes this loop.

Applying the DMD generator objective from Self-Forcing~\citep{huang2025selfforcing}, we perturb the rollout prediction with fresh noise, \(\hat{\mathbf{x}}_t=(1-t)\hat{\mathbf{x}}_0+t\epsilon\) for \(\epsilon\sim\mathcal{N}(0,I)\), and train the student with
\begin{equation}
\mathcal{L}_{\mathrm{sf}}=\mathbb{E}_{t,\epsilon}\left[\frac{1}{2}\left\|\hat{\mathbf{x}}_0-\mathrm{sg}\left(\hat{\mathbf{x}}_0-\left(s_{\mathrm{fake}}(\hat{\mathbf{x}}_t,c,t)-s_{\mathrm{real}}(\hat{\mathbf{x}}_t,c,t)\right)\right)\right\|_2^2\right].
\end{equation}
Here \(\mathrm{sg}(\cdot)\) denotes stop gradient, \(s_{\mathrm{real}}\) is a frozen score network initialized from the Stage 1 teacher, and \(s_{\mathrm{fake}}\) is a trainable critic on student rollouts. Because frame \(\mathbf{x}_i\) commits \(\mathbf{r}_i\) during rollout, Stage 3 reuses the same register loss on registers that encode world states of completed context frames:
\[
\mathcal{L}_{\mathrm{S3}}=\mathcal{L}_{\mathrm{sf}}+\lambda_{\mathrm{state}}\mathcal{L}_{\mathrm{reg}}.
\]

We likewise set \(\lambda_{\mathrm{state}}\) to one by default in Stage 3.
In practice, we follow Solaris~\citep{savva2026solaris} by using a four step denoising schedule with timesteps \(1000\to750\to500\to250\to0\) and by sampling an early exit denoising step during rollout for stochastic gradient truncation.
At this stage, we find that the DMD loss alone is not sufficient to improve stability, since WSR modeling also suffers from drift.
Thus, we increase selected per-head register supervision weights relative to Stage 2, which helps train better registers and stabilize long horizon rollout.


\section{Experiments}
\label{sec:experiments}


In this section, we first present the main results in Sec.~\ref{sec:main-results}, then analyze the contribution of each supervision signal in Sec.~\ref{sec:supervision-ablation}, study the model architecture in Sec.~\ref{sec:exp-model-arch}, and discuss semi-supervised training in Sec.~\ref{sec:semi-sup}.

\subsection{Experimental Setup}
\label{sec:exp-setup}

\smallskip
\noindent
\textbf{Training data.} We build on our data collection pipeline based on SolarisEngine~\citep{savva2026solaris}. We collect approximately 126 hours of synchronized two-player videos with extensive agent status information and additional bird's-eye view camera footage. We further annotate the scene text of each 4-frame clip with Qwen2.5-VL-72B-Instruct~\citep{bai2025qwen25vltechnicalreport}. Detailed data collection and annotation are provided in Appendix~\ref{app:data}.

\smallskip
\noindent
\textbf{Evaluation.} We evaluate our model on the Solaris original test split~\citep{savva2026solaris}. For each clip, the model receives both players' first-frame observations and full action sequences, then generates the remaining rollout. We report VLM accuracy and FID for each evaluation category. VLM accuracy measures whether the generated video preserves the queried world relation, such as relative object position and cross-player consistency, while FID measures visual realism and distributional fidelity. To summarize these two axes without replacing either, we also report an auxiliary \textbf{world score}:
{\setlength{\abovedisplayskip}{4pt}\setlength{\belowdisplayskip}{4pt}
\begin{equation}
\mathrm{WorldScore}=
\frac{\frac{1}{K}\sum_{k=1}^{K}\mathrm{VLM}_{k}}
{\frac{1}{K}\sum_{k=1}^{K}\mathrm{FID}_{k}},
\end{equation}
}
where \(K=5\). We report VLM accuracy in percentage points and multiply the ratio by 100 for readability. Since WorldScore is unbounded and sensitive to the scale of FID, we use it only as an auxiliary ranking metric. VLM accuracy and FID remain the primary metrics.

\subsection{Main Results}
\label{sec:main-results}

We compare our \ourmodel multi-agent world model with the following baselines:
(1) \emph{Frame concat}: a world model that concatenates the two players' observations along the channel dimension, following Multiverse~\citep{enigma2025multiverse};
(2) \emph{Solaris}~\citep{savva2026solaris}: a multi-agent world model that synchronizes multi-agent observations by extending the sequence of frame tokens and using bidirectional attention across players. To ensure a fair comparison, we retrain the Solaris model with the same training data as our model.

As shown in Table~\ref{tab:main}, \ourmodel improves the aggregate world score to 105.1, surpassing the previous best by a large margin.
The VLM accuracy gains are strongest on state-sensitive categories. Compared with the Solaris baseline, Grounding rises from 81.3 to 93.8, Building rises from 9.4 to 28.1, and Consistency rises from 57.8 to 76.6. These gains suggest that the persistent registers do more than improve visual fidelity: they help the model maintain a logically coherent and verifiable world model across players and rollout steps.

\begin{figure*}[t]
    \centering
    \includegraphics[width=\textwidth]{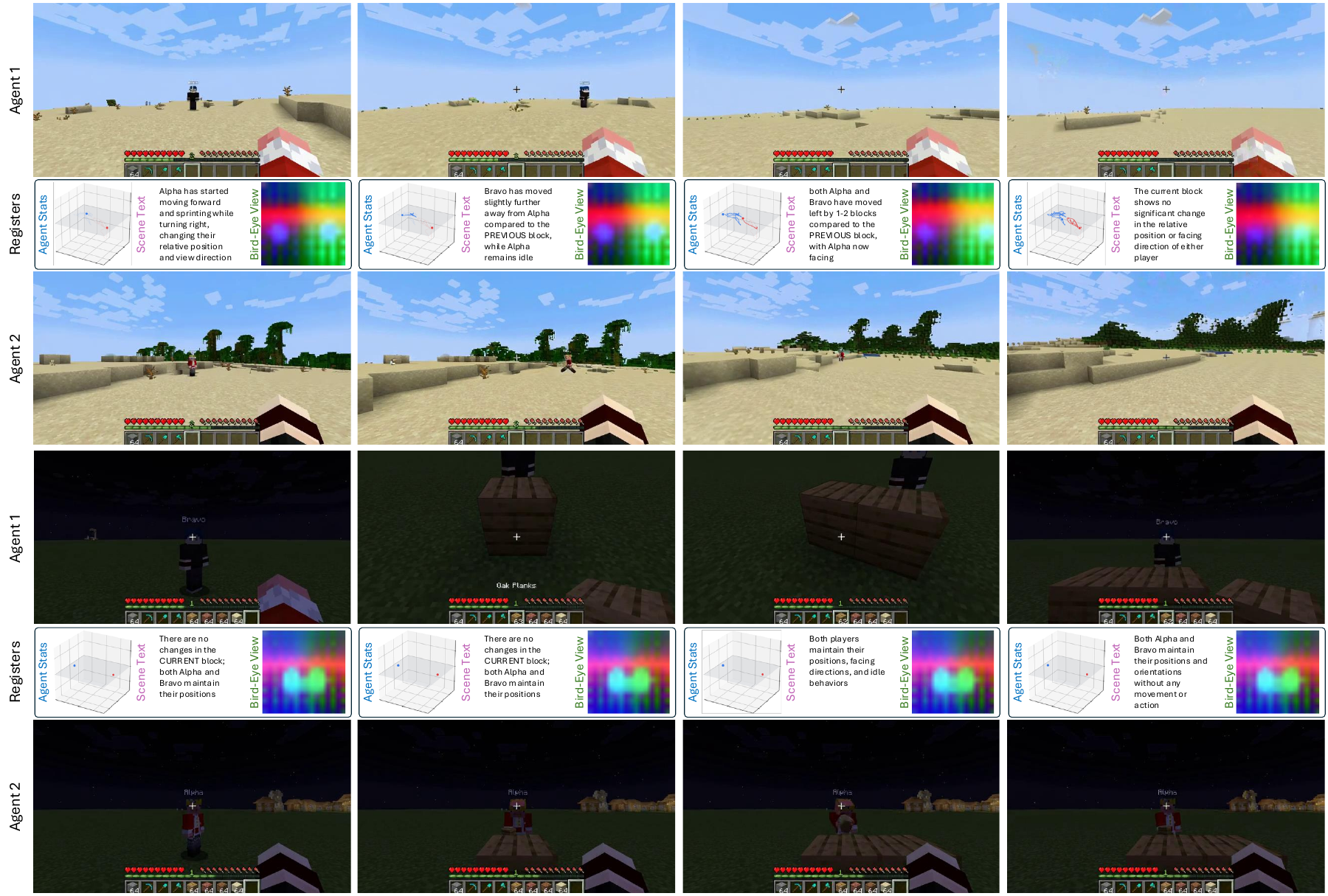}
    \vspace{-1.5em}
    \caption{\textbf{Qualitative visualization.} We show generated multi-agent rollouts with grounded world states: agent coordinates, bird's-eye-view maps, and scene text. These signals make the persistent world state inspectable and help verify cross-player consistency over time. Please refer to the project page for more detailed video demo.
    }
    \vspace{-1.0em}
    \label{fig:qualitative-visualization}
\end{figure*}

Figure~\ref{fig:qualitative-visualization} visualizes the generated world together with decoded world states, including agent statistics, bird's-eye-view maps, and scene text. For agent statistics, we selectively visualize the positions and trajectories. For the bird's-eye-view maps, we use PCA to project the predicted DINOv2~\citep{oquab2023dinov2} features to RGB images.
WSR rollout allows \ourmodel to generate a logically coherent world and better synchronize cross-agent information.
Additionally, WSR offers a more interpretable way to inspect the state represented by each generated chunk.

\begin{table}[t]
\centering
\small
\caption{Comparison with multi-agent world model baselines. We report category-level VLM accuracy, FID~\citep{heusel2017fid}, and the aggregate world score. Best result in each column is highlighted in \textbf{bold}.}
\label{tab:main}
\makebox[\linewidth][c]{%
\resizebox{0.98\linewidth}{!}{%
\begin{tabular}{lcccccccccc@{\hspace{5pt}}|@{\hspace{5pt}}c}
\toprule
 & \multicolumn{2}{c}{Movement} & \multicolumn{2}{c}{Grounding} & \multicolumn{2}{c}{Memory} & \multicolumn{2}{c}{Building} & \multicolumn{2}{c@{\hspace{5pt}}|@{\hspace{5pt}}}{Consistency} & World \\
\cmidrule(lr){2-3} \cmidrule(lr){4-5} \cmidrule(lr){6-7} \cmidrule(lr){8-9} \cmidrule(lr){10-11}
Method & VLM $\uparrow$ & FID $\downarrow$ & VLM $\uparrow$ & FID $\downarrow$ & VLM $\uparrow$ & FID $\downarrow$ & VLM $\uparrow$ & FID $\downarrow$ & VLM $\uparrow$ & FID $\downarrow$ & Score $\uparrow$ \\
\midrule
Frame concat                         & 77.1 & 68.9 & 53.1 & 66.6 & 37.5 & 74.4 & 0.0 & 103.2 & 49.5 & 129.4 & 49.1 \\
Solaris$^{*}$ & 79.7 & 43.3 & 81.3 & 37.2 & 43.8 & \textbf{61.2} & 9.4 & 83.4 & 57.8 & 110.7 & 81.0 \\
\midrule
\textbf{\ourmodelshort{} (Ours)}                    & \textbf{82.8} & \textbf{34.0} & \textbf{93.8} & \textbf{36.8} & \textbf{46.9} & 64.8 & \textbf{28.1} & \textbf{75.9} & \textbf{76.6} & \textbf{100.7} & \textbf{105.1} \\
\bottomrule
\end{tabular}%
}%
}
\end{table}

\subsection{Analysis of Supervision Signals}
\label{sec:supervision-ablation}

A core problem is not only how to supervise the world state registers, but also why explicit supervision is needed. 
Ideally, the registers could learn both agent-specific state and global scene information from the diffusion loss alone. 
However, in practice, this signal does not specify which information should be stored as world state.
We therefore add complementary supervision signals that encourage the registers to encode meaningful and useful world state information. 


We hypothesize that the world state registers should encode local dynamics, global geometry, and cross-modal semantics.
We therefore instantiate the three supervision types as follows.
(1) For individual-state supervision, the register predicts \emph{agent statistics}, including each player's position, velocity, and orientation.
These targets give the register an explicit per-agent coordinate signal, allowing us to test whether the persistent state carries verifiable motion information for each player.
(2) For global 3D supervision, we use a \emph{bird's-eye view} that exposes scene layout and object distribution from an allocentric viewpoint.
This signal asks the register to preserve spatial structure shared by both players, rather than only the egocentric evidence visible in the current frame.
(3) For global cross-modal supervision, we attach \emph{scene text} that describes the visual state of the current frame in language.
This target complements local dynamics and global geometry, asking the register to retain categories, attributes, and high-level content across rollouts.

\begin{table}[t]
\centering
\small
\caption{Supervision ablation for world state registers. We compare registers without explicit targets and registers with different signals; their combination further improves the world state registers.
}
\label{tab:supervision-ablation}
\makebox[\linewidth][c]{%
\resizebox{0.98\linewidth}{!}{%
\begin{tabular}{lcccccccccc@{\hspace{5pt}}|@{\hspace{5pt}}c}
\toprule
 & \multicolumn{2}{c}{Movement} & \multicolumn{2}{c}{Grounding} & \multicolumn{2}{c}{Memory} & \multicolumn{2}{c}{Building} & \multicolumn{2}{c@{\hspace{5pt}}|@{\hspace{5pt}}}{Consistency} & World \\
\cmidrule(lr){2-3} \cmidrule(lr){4-5} \cmidrule(lr){6-7} \cmidrule(lr){8-9} \cmidrule(lr){10-11}
Variant & VLM $\uparrow$ & FID $\downarrow$ & VLM $\uparrow$ & FID $\downarrow$ & VLM $\uparrow$ & FID $\downarrow$ & VLM $\uparrow$ & FID $\downarrow$ & VLM $\uparrow$ & FID $\downarrow$ & Score $\uparrow$ \\
\midrule
Baseline                 & 79.7 & 43.3 & 81.3 & 37.2 & 43.8 & 61.2 & 9.4 & 83.4 & 57.8 & 110.7 & 81.0 \\
Registers only           & 90.6 & 43.2 & 81.3 & 44.3 & \textbf{62.5} & 66.1 & 21.9 & 84.3 & 62.5 & 101.9 & 93.8 \\
~ + Agent stats  & \textbf{95.3} & 41.0 & 59.4 & 45.5 & 56.3 & \textbf{60.1} & 9.4 & 80.8 & 75.0 & 107.9 & 88.1 \\
~ + Bird's-eye view    & 82.8 & 39.1 & \textbf{96.9} & 40.7 & 46.9 & 64.7 & \textbf{31.3} & \textbf{74.2} & 71.9 & 103.4 & 102.4 \\
~ + Scene text       & 85.9 & 40.2 & 84.4 & 38.4 & \textbf{62.5} & 62.1 & 25.0 & 78.8 & 73.4 & 101.6 & 103.2 \\
~ + All              & 82.8 & \textbf{34.0} & 93.8 & \textbf{36.8} & 46.9 & 64.8 & 28.1 & 75.9 & \textbf{76.6} & \textbf{100.7} & \textbf{105.1} \\
\bottomrule
\end{tabular}%
}%
}
\vspace{-1.5em}
\end{table}

Starting from the Solaris~\citep{savva2026solaris} baseline, we first enable registers without explicit supervision, then add each signal, and finally evaluate the combined setting, as reported in Table~\ref{tab:supervision-ablation}. All other hyperparameters stay at the defaults of Section~\ref{sec:exp-setup}.
First, we observe that registers alone already help: without any explicit state target, world score improves from 81.0 to 93.8.
We attribute this gain to registers implicitly building a persistent world state, because they provide a dedicated slot to carry cross-agent information instead of recomputing it from the local window at every step.

However, explicitly modeling world-state targets can push performance further compared with the baseline.
All three supervised variants outperform the baseline (81.0) because they specify what the registers should encode and keep verifiable across rollout steps, instead of leaving that choice entirely to the pixel loss.
We find that scene text helps the most among single supervision signals, raising world score to 103.2.
Combining all three signals further lifts world score from 93.8 to 105.1, with higher mean VLM accuracy and lower mean FID.
These results corroborate our hypothesis and highlight the importance of enabling world state registers in the multi-agent world model setting.

\subsection{Analysis of Model Architecture}
\label{sec:exp-model-arch}

Motivated by recent studies in multimodal generation~\citep{mo2025xfusion}, we view the architecture as a possible source of conflict between different token roles.
A dense transformer can in principle use the same parameters to render pixel context frames and generate accurate world state registers.
However, register tokens are intended to represent hidden information about the world, while frame tokens mainly carry local visual information.
This difference in function becomes harder to balance once the register encodes richer state information.
We follow the MoT design by using separate modality-specific weights to process register and frame tokens, while allowing them to interact through the shared attention mechanism.


Table~\ref{tab:mot-config} compares dense and MoT backbones under two settings, using registers without explicit supervision and using registers with scene text supervision. The results suggest that MoT is not automatically beneficial. Without explicit supervision, MoT does not improve the aggregate world score over the dense backbone. 
The pattern changes under scene text supervision: the dense model drops to 91.3 world score, while MoT reaches 103.2. This contrast suggests that richer semantic representations in world state registers are harder to process with the same parameterization as visual tokens. Separating register and frame computation therefore becomes useful when the register carries a distinct semantic world state.

\begin{wrapfigure}{r}{0.35\linewidth}
\centering
\vspace{-1.5em}
\includegraphics[width=\linewidth]{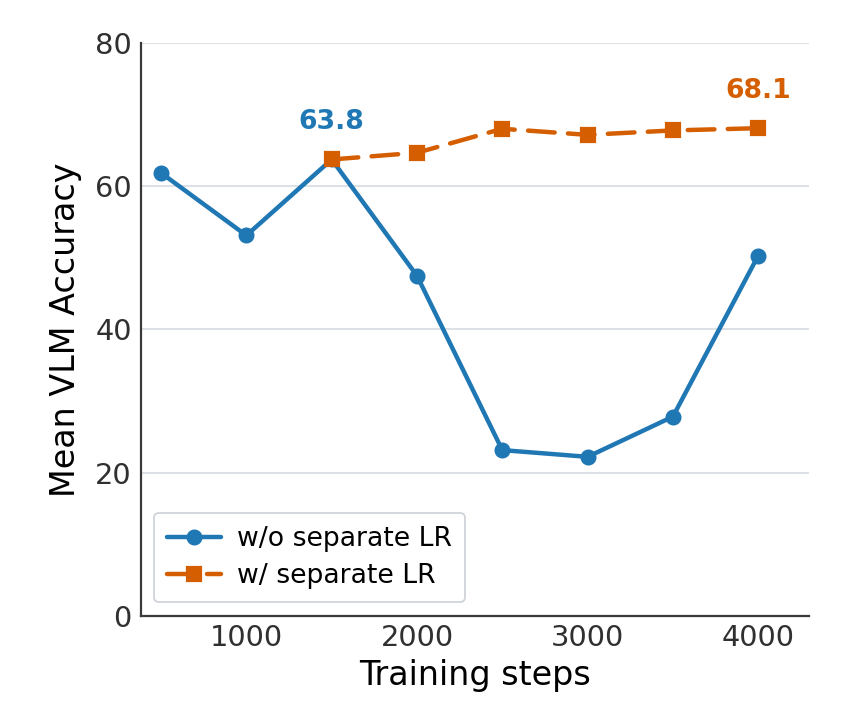}
\vspace{-2.0em}
\caption{\textbf{VLM accuracy over Stage 3 self-forcing.}
}
\vspace{-2.0em}
\label{fig:score-vs-steps}
\end{wrapfigure}
Additionally, we investigate the training dynamics of the MoT model during Stage 3 training.
Separate weight parameters also allow a more controlled learning process.
We empirically find that the register loss in Stage 3 training converges more slowly than the DMD distillation loss, while the register loss mainly updates the register transformer and the DMD distillation loss mainly updates the visual transformer.
Therefore, after step 1500, we apply the separate LR schedule, where we freeze the weight parameters of the visual transformer and only update the weight parameters of the register transformer during Stage 3 training.
As shown in Figure~\ref{fig:score-vs-steps}, the mean VLM accuracy improves from 63.8 to 68.1.
This result suggests that the MoT is important for WSR-based multi-agent world modeling.

\begin{table}[t]
\centering
\small
\caption{Dense and MoT backbone ablation. We compare a dense transformer with a Mixture of Transformers under no register supervision and scene text supervision. }
\label{tab:mot-config}
\makebox[\linewidth][c]{%
\resizebox{0.98\linewidth}{!}{%
\begin{tabular}{lcccccccccc@{\hspace{5pt}}|@{\hspace{5pt}}c}
\toprule
 & \multicolumn{2}{c}{Movement} & \multicolumn{2}{c}{Grounding} & \multicolumn{2}{c}{Memory} & \multicolumn{2}{c}{Building} & \multicolumn{2}{c@{\hspace{5pt}}|@{\hspace{5pt}}}{Consistency} & World \\
\cmidrule(lr){2-3} \cmidrule(lr){4-5} \cmidrule(lr){6-7} \cmidrule(lr){8-9} \cmidrule(lr){10-11}
Model & VLM $\uparrow$ & FID $\downarrow$ & VLM $\uparrow$ & FID $\downarrow$ & VLM $\uparrow$ & FID $\downarrow$ & VLM $\uparrow$ & FID $\downarrow$ & VLM $\uparrow$ & FID $\downarrow$ & Score $\uparrow$ \\
\midrule
\multicolumn{12}{l}{\emph{No supervision}} \\
Dense & 76.6 & 41.7 & 87.5 & 40.4 & 65.6 & 54.0 & 12.5 & 87.2 & 73.4 & 107.4 & 95.5 \\
MoT   & 90.6 & 43.2 & 81.3 & 44.3 & 62.5 & 66.1 & 21.9 & 84.3 & 62.5 & 101.9 & 93.8 \\
\midrule
\multicolumn{12}{l}{\emph{Scene text supervision}} \\
Dense & 81.3 & 37.3 & 81.3 & 53.1 & 53.1 & 54.8 & 9.4 & 73.9 & 71.9 & 106.4 & 91.3 \\
MoT   & 85.9 & 40.2 & 84.4 & 38.4 & 62.5 & 62.1 & 25.0 & 78.8 & 73.4 & 101.6 & 103.2 \\
\bottomrule
\end{tabular}%
}%
}
\vspace{-1.0em}
\end{table}



\subsection{Extension: Semi-Supervised Training}
\label{sec:semi-sup}

A natural question is whether this pipeline can extend beyond fully instrumented simulator data toward more realistic world modeling. 
It remains challenging to collect real-world data with explicit world-state annotations \eg agent stats and bird's-eye-view maps.
We therefore adopt a semi-supervised setup: clips with bird's-eye-view maps are labeled, and clips without them are unlabeled.
This setup serves as a proxy for future realistic data regimes, where raw videos may be abundant but explicit world-state labels are limited.

\begin{table}[t]
\centering
\small
\caption{Semi-supervised training. Each row varies the number of data clips without bird's-eye view annotation (unlabeled data), where ``1K / N~K'' denotes 1K labeled clips and N~K unlabeled clips.}
\label{tab:semi-sup}
\makebox[\linewidth][c]{%
\resizebox{0.98\linewidth}{!}{%
\begin{tabular}{lcccccccccc@{\hspace{5pt}}|@{\hspace{5pt}}c}
\toprule
 & \multicolumn{2}{c}{Movement} & \multicolumn{2}{c}{Grounding} & \multicolumn{2}{c}{Memory} & \multicolumn{2}{c}{Building} & \multicolumn{2}{c@{\hspace{5pt}}|@{\hspace{5pt}}}{Consistency} & World \\
\cmidrule(lr){2-3} \cmidrule(lr){4-5} \cmidrule(lr){6-7} \cmidrule(lr){8-9} \cmidrule(lr){10-11}
Data Split & VLM $\uparrow$ & FID $\downarrow$ & VLM $\uparrow$ & FID $\downarrow$ & VLM $\uparrow$ & FID $\downarrow$ & VLM $\uparrow$ & FID $\downarrow$ & VLM $\uparrow$ & FID $\downarrow$ & Score $\uparrow$ \\
\midrule
1K / 0K    & 78.1 & 60.0 & 75.0 & 72.6 & 21.9 & 95.0 & 15.6 & 76.9 & 71.9 & 111.0 & 63.2 \\
1K / 2.5K  & 68.8 & 58.8 & 56.3 & 60.2 & 65.6 & 96.4 & 15.6 & 79.7 & 56.3 & 112.4 & 64.4 \\
1K / 5.0K  & 85.9 & 60.4 & 56.3 & 58.3 & 81.3 & 79.8 & 28.1 & 76.6 & 71.9 & 117.7 & 82.3 \\
1K / 10.0K & 81.3& 44.3& 65.6& 53.2& 84.3& 75.8& 25.0& 73.6& 68.8& 112.9& 90.3\\
\bottomrule
\end{tabular}%
}%
}
\end{table}

In our experimental setting, we fix the labeled set to 1K video clips and gradually increase the amount of unlabeled video clips.
Labeled samples receive both the diffusion loss and the register supervision losses, while unlabeled samples train the video generator through the diffusion objective only. 
We apply bird's-eye view supervision and use the default training hyperparameters except for the Stage-2 training length, where we use 20K instead of 60K steps to prevent overfitting to the labeled data.
Table~\ref{tab:semi-sup} shows that adding unlabeled data improves the aggregate world score from 63.2 with no unlabeled data to 82.3 with 5K unlabeled clips, and further to 90.3 with 10K unlabeled clips. 
We observe a generally consistent improvement in both VLM accuracy and FID across the table, suggesting that unlabeled rollout data can still improve generation quality when a small supervised subset anchors the hidden world state in the registers.
Overall, our approach provides a promising direction for settings where explicit world-state labels are hard to collect.

\section{Conclusion}
\label{sec:conclusion}

We introduce \ourmodel{}, a streaming multi-agent video diffusion model with cross-agent world state registers for interactive video generation.
The core idea is to maintain and update a shared world state across agents.
Our model then refines remembered scene information, rather than repeatedly relying on an expanding frame history.
Across multi-agent Minecraft experiments, this design improves both visual quality and logical consistency, and the ablations show that explicit world-state supervision further shapes the registers into a more verifiable state representation.

\textbf{Limitations.}
While our experiments highlight the potential of cross-agent world state registers, there are still several limitations.
Our key improvement comes from the additional state supervision, but the state in the real world is much more complex and is not directly available.
However, we believe this is a promising direction for future work, where a world model could be aware of more complex state information, e.g., from the low-level 3D consistent visual detail to the high-level semantic relationships across agents and players.
We hope this work could shed light on future work on world models with cross-agent capabilities.


{
    \small
    \bibliographystyle{iclr2026_conference}
    \bibliography{main}
}

\newpage

\appendix
\appendixnavigation

\section{Data Collection}
\label{app:data}
\label{app:data-collection}

\subsection{Overview}

We build our dataset on top of the SolarisEngine simulator~\citep{savva2026solaris}. The dataset contains approximately 126 hours of synchronized two-agent interaction recordings at a resolution of $832\times480$ and a frame rate of 20 fps. Each recording session pairs two agents, denoted Alpha and Bravo, acting concurrently in the same episode, and each agent's state is logged at 20 Hz alongside the video stream. Every session is paired with three complementary streams of world-state annotations: per-agent statistics that capture individual dynamics, a shared bird's-eye view that exposes the global scene layout, and per-chunk scene text that provides cross-modal semantics. These signals supply the individual, global, and cross-modal supervision used to ground the world-state registers, and we detail how each stream is collected in the following subsections.

\subsection{Individual states: agent statistics}
For the individual state stream, we log each agent's position, velocity, and orientation, along with ground-truth controller inputs at 20 Hz, giving a compact state record that captures how each agent moves through the episode. Figure~\ref{fig:data-vis} visualizes these statistics for Alpha and Bravo over time.

\subsection{Global state: bird's-eye view}
In addition to each agent's egocentric camera, every recording session is filmed by a shared fixed overhead camera positioned above the scene. Because this camera is shared rather than attached to either agent, both agents appear together in a single frame, giving a global view of their relative position and spatial relationship that complements the two first-person streams. Figure~\ref{fig:data-vis} shows representative frames from this overhead camera over the course of one episode.

\subsection{Cross-modal states: scene text}
The scene text collection process is separate from the game engine. To provide language-grounded supervision for cross-modal scene understanding, we annotate every recording with Qwen2.5-VL-72B-Instruct. Each episode is divided into temporal chunks of four consecutive frames, corresponding to 200 ms at 20 fps and matching the temporal downsampling factor of our video tokenizer. For each chunk, we assemble synchronized inputs from the shared overhead view and the two egocentric views. We sample three frames from each view for the current chunk, and we include the immediately preceding chunk as temporal context when it is available. Generation is capped at 256 tokens.

To ground the caption in observed behavior rather than asking the model to infer actions from pixels alone, we additionally condition the prompt on each agent's ground-truth controller input summary for the current chunk, including movement keys, action keys, hotbar or tool selection, and net camera rotation. We use the following question template when captioning chunk $t$ and provide chunk $t-1$ as temporal context whenever it exists:

\begin{quote}\small
Synchronized short blocks of a two-agent session are shown in time order: first the previous block, then the current block. For each block you see an overhead view, then Alpha's egocentric view, then Bravo's egocentric view. Ignore corner text, chat, heart and hunger bars; the bottom bar in an egocentric view is that player's hotbar.

Ground-truth controller inputs for the current block: Alpha [current input summary], Bravo [current input summary]. Ground-truth controller inputs for the previous block: Alpha [previous input summary], Bravo [previous input summary].

Using both the views and these inputs, fill in this template exactly, one sentence per line:

Overall: what each agent does in the current block, their relative position and distance, and where each is facing or looking.

Change: how the current block differs from the previous one in relative position, facing direction, view direction, and each agent's behavior.
\end{quote}

For an episode's first chunk, which has no previous chunk, we omit the temporal context and use a shorter prompt whose change line is fixed to ``first block, no previous, none.''

\begin{figure}[t]
    \centering
    \includegraphics[width=\linewidth]{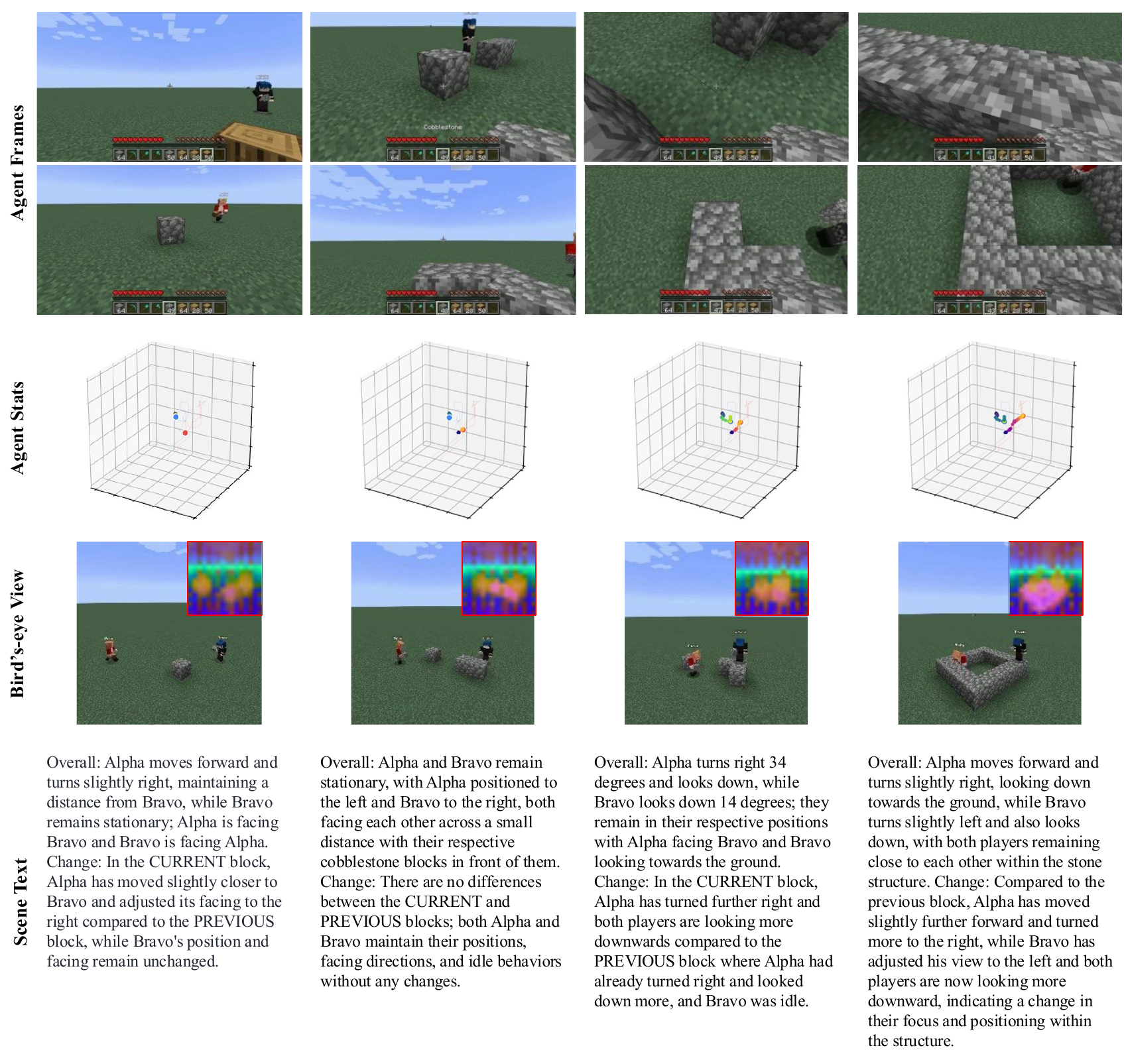}
    \caption{Visualization of the data collection and annotation streams used in our multi-agent Minecraft setting.}
    \label{fig:data-vis}
    \end{figure}

Figure~\ref{fig:data-vis} shows representative chunk inputs alongside their generated captions. Because this pipeline is built around each agent's local motion and controller inputs, the resulting captions describe per-agent behavior faithfully but do not capture broader scene context beyond the two agents, such as surrounding terrain or events outside either agent's immediate vicinity.

\clearpage
\section{Pipeline and implementation details}
\label{app:pipeline}

\ourmodel{} turns a single player video diffusion prior into a streaming two player world model by adding an explicit persistent state pathway to the autoregressive rollout. The pipeline has three implementation components. First, committed world state registers are decoded during training with auxiliary heads for agent statistics, bird's-eye view features, and scene text. Second, a Mixture of Transformers backbone routes register tokens and player frame tokens through role specific parameter branches while preserving joint attention over the interleaved sequence. Third, training follows a three stage curriculum that first adapts the video prior to synchronized two player data, then trains a causal register student, and finally exposes the student to its own generated frames and committed registers under self-forcing.

\subsection{State decoding}

\begin{table}[h]
\centering
\small
\setlength{\tabcolsep}{8pt}
\caption{Configuration of the three world state register supervision signals. Each signal has its own training only prediction head \(h_m\), target \(\mathbf{y}^m\), distance metric \(d_m\), and loss weight \(\lambda_m\). The combined setting activates all heads jointly and sums the per signal losses. Heads are discarded at inference.}
\label{tab:supervision-hparams}
\begin{tabular}{lcccc}
\toprule
 & Agent states & Bird's-eye view & Scene text & Combined \\
\midrule
\multicolumn{5}{l}{\textbf{Decoder Architecture}} \\
Layers                & 4 & 12 & 16 & - \\
Channels              & 256 & 768 & 2048 & - \\
Query Number          & 2 & 256 & - & - \\
WSR number            & 256 & 256 & 256 & 64/64/128 \\
Weight Init            & Random & Random & Pretrained & - \\
Learnable             & True & True & False & - \\
\midrule
\multicolumn{5}{l}{\textbf{Optimization}} \\
Distance metric \(d_m\)   & MSE & Cos. Sim. & Cross Entropy & - \\
S2: Loss weight \(\lambda_m\) & 0.02 & 2.0 & 0.1 & 0.02/2.0/0.1 \\
S3: Loss weight \(\lambda_m\) & 0.2 & 2.0 & 0.5 & 0.2/2.0/0.5 \\


\bottomrule
\end{tabular}%
\end{table}

Table~\ref{tab:supervision-hparams} summarizes the numeric configuration, and this subsection provides a more careful description of the state decoding supervision used to train the world state registers. The common pattern is to project committed world state register (WSR) tokens into the decoder space and place them in an attention sequence together with the tokens that will receive supervision. For the transformer based heads, this sequence is processed by full self attention; for scene text, the projected registers play the same role as prefix embeddings for the frozen language model. In this view, the projected registers serve as prefix like conditioning tokens: the remaining query, patch, or caption tokens attend to this state representation and are then scored by the corresponding target. The following paragraphs first describe this shared decoding pattern, then give the concrete targets and losses for agent statistics, bird's-eye view features, and scene text.

The objective in the main text already defines the weighted register loss, so we focus here on implementation details that are not explicit in the main description. Each head decodes every committed register step, with no pooling over WSR tokens. Single signal ablations may read the full register bank, whereas the combined setting uses disjoint \(64/64/128\) slices. Missing targets are skipped for the affected head, and the implementation uses the per head weights in Table~\ref{tab:supervision-hparams} directly, with no separate global state weight.

\subsubsection{Agent statistics decoding}

The appendix specifies the state vector used by the agent head. Each player is represented by position, velocity, and orientation, matching the agent-state target in the main text. Targets are selected at the latent aligned simulator frame and kept in raw units. We do not apply per field normalization, clipping, or temporal smoothing.

\subsubsection{Bird's-eye view decoding}

The bird's-eye view signal is feature supervision rather than pixel reconstruction. The decoder predicts a \(16{\times}16\) grid of 768 dimensional DINOv2 ViT-B/14 patch features from 256 learnable patch queries. The target top down frame is processed by the data pipeline and then by the frozen DINOv2 preprocessing path to \(224{\times}224\); we use dense patch tokens rather than the CLS token or a pooled feature. PCA to RGB is used only for qualitative visualization.

\subsubsection{Scene text decoding}

The scene text head should not be read as a direct vocabulary logit head. The trainable module projects the selected WSR slice into 2048 dimensional prefix embeddings for a frozen Llama-3.2-1B~\citep{grattafiori2024llama} scorer; the frozen language model, not the trainable head, produces the next token logits. Captions are tokenized to 128 training slots, with padding and BOS ignored in the cross entropy. The caption targets are authored offline by Qwen2.5-VL-72B-Instruct~\citep{bai2025qwen25vltechnicalreport}, which is separate from the frozen Llama model used for supervision.

\subsection{Mixture of Transformers}
\label{app:mot-implementation}

Alongside the world state registers, we introduce a Mixture of Transformers (MoT) backbone into the autoregressive diffusion design.
Importantly, MoT changes only how the generator's parameters are shared while the streaming rollout, the interleaved token layout, and the causal KV-cache all stay the same.
Concretely, WSR tokens use the state branch, while the content frame tokens use the visual branch. In our implementation, one branch includes the transformer projection layers, feed forward, conditioning, and normalization layers. We initialize the state branch weights using the pretrained visual branch weights.
It is worth noting that for self attention, although the register tokens and content frame tokens are encoded by different projection matrices, the attention weights are computed jointly over the interleaved sequence.
Since the branches are shape matched, every token still activates only a single parameter copy, so per-token FLOPs are unchanged apart from routing overhead even as the total parameter count grows.


\subsection{Training}

\begin{table}[h]
\centering
\small
\setlength{\tabcolsep}{8pt}
\caption{Hyperparameter setup for the three stage training curriculum used by the primary combined world state model. All stages use \(P{=}2\) players. Entries marked ``--'' do not apply to that stage.}
\label{tab:hyperparams}
\begin{tabular}{lccc}
\toprule
 & Stage 1 & Stage 2 & Stage 3 \\
& Bidirectional & Causal Training & Self-Forcing \\
\midrule
\multicolumn{4}{l}{\textbf{Architecture}} \\

Sliding window \(W\)            & -- & 6 & 6 \\
Register tokens \(K\)           & -- & 256 & 256 \\
Backbone                        & Dense & MoT & MoT \\
\midrule
\multicolumn{4}{l}{\textbf{Optimization}} \\
Training steps                  & 120K & 80K & 3-4K \\
Batch size                      & 32 & 32 & 32 \\
Optimizer                       & AdamW & AdamW & AdamW \\
\(\beta\)                       & \((0.9,0.95)\) & \((0.9,0.95)\) & \((0.9,0.95)\) \\
Learning rate                   & \(1{\times}10^{-4}\) & \(1{\times}10^{-4}\) & \(3{\times}10^{-6}\) \\
\bottomrule
\end{tabular}%
\end{table}

Table~\ref{tab:hyperparams} lists the hyperparameters of the primary combined setting.
Shared choices are \(P{=}2\), \(K{=}256\) registers with the \(64/64/128\) head grouping, \(W{=}6\) in causal stages, AdamW with bf16 mixed precision, and a batch size of 32.

\subsubsection{Stage 1: Bidirectional Training}
We initialize from the official Solaris bidirectional checkpoint~\cite{savva2026solaris}.
A player axis is added to observations, actions, VAE latents, and first-frame conditioning, and no register pathway is used.
Training runs for 120K steps under Table~\ref{tab:hyperparams}.

\subsubsection{Stage 2: Causal Training with World State Registers}
We first train a causal student without registers for 60K steps, then continue with WSR, MoT, and the three state heads for 20K steps.
This 60K{+}20K schedule matches training with WSR for the full 80K steps in our checks.
Following Solaris~\citep{savva2026solaris},we train this model with flow matching loss and register loss on ground-truth latents under Diffusion Forcing noise, with no live teacher or ODE rollout.
Register weights follow Table~\ref{tab:supervision-hparams} with no extra global multiplier.

\subsubsection{Stage 3: Self-Forcing with Context Frame and State Rollout}
The generator loads the matching Stage-2 combined checkpoint.
Both \(s_{\mathrm{real}}\) and \(s_{\mathrm{fake}}\) start from the Stage-1 bidirectional checkpoint; \(s_{\mathrm{real}}\) stays frozen and \(s_{\mathrm{fake}}\) is trained separately.
Each update runs a no-gradient rollout on \(1000{\to}750{\to}500{\to}250{\to}0\) with a shared early exit, then a gradient pass on the generated context for the DMD and register losses.
Relative to Stage 2, agent-statistics and scene-text weights are raised and the BEV weight is unchanged (Table~\ref{tab:supervision-hparams}).
Training lasts 3--4K steps.
The MoT freeze schedule in Section~\ref{sec:exp-model-arch} is a separate continuation that locks all non-register generator parameter groups.



\end{document}